\documentclass[10pt,twocolumn,letterpaper]{article}

\usepackage[pagenumbers]{cvpr} 
\usepackage{color}
\usepackage{colortbl}
\usepackage{multirow}
\usepackage{algorithm}
\usepackage{algorithmicx}
\usepackage{algpseudocode}
\usepackage{amsmath}
\usepackage{graphicx}
\usepackage{booktabs}

\definecolor{cvprblue}{rgb}{0.21,0.49,0.74}
\definecolor{lightgray}{gray}{0.95}
\usepackage[pagebackref,breaklinks,colorlinks,allcolors=cvprblue]{hyperref}

\def\paperID{*****}
\def\confName{CVPR}
\def\confYear{2025}

\title{GridPrune: \\ From ``Where to Look" to ``What to Select" in Visual Token Pruning for MLLMs}

\author{
Yuxiang Duan\textsuperscript{1} \quad 
Ao Li\textsuperscript{1} \quad 
Yingqin Li\textsuperscript{1} \quad
Luyu Li\textsuperscript{1} \quad
Pengwei Wang\thanks{Corresponding author.}\textsuperscript{\rm ~~,1}
\\
\textsuperscript{1}Shandong University~
}

\begin{document}
\maketitle
\begin{abstract}
Multimodal large language models (MLLMs) have shown remarkable capabilities in a wide range of vision-language tasks. 
However, the large number of visual tokens introduces significant computational overhead.
To address this issue, visual token pruning has emerged as a key technique for enhancing the efficiency of MLLMs.
In cognitive science, humans tend to first determine which regions of a scene to attend to (``where to look") before deciding which specific elements within those regions to process in detail (``what to select").
This two-stage strategy enables the visual system to efficiently allocate attention at a coarse spatial level before performing fine-grained selection.
However, existing pruning methods primarily focus on directly optimizing ``what to select", typically using attention scores or similarity metrics.
They rarely consider ``where to look", which has been shown to lead to inefficient spatial allocation, positional bias, and the retention of irrelevant or redundant tokens.
In this paper, we propose \textbf{GridPrune}, a method that replaces the global Top-K mechanism with a ``guide-globally, select-locally" zonal selection system.
GridPrune splits the pruning process into two steps: first, it uses text-conditional guidance to dynamically allocate a token budget across spatial zones;
and then, it performs local selection within each budgeted zone.
Experimental results demonstrate that GridPrune achieves superior performance across various MLLM architectures.
On LLaVA-NeXT-7B, GridPrune retains 96.98\% of the full performance while using 11.1\% of the tokens, outperforming the best-performing baseline by 2.34\% at the same pruning rate.
\end{abstract}

\section{Introduction}

Multimodal Large Language Models (MLLMs)\cite{llava,llava1.5,qwen2.5vl,internvl,instructblip,minigpt4} have shown remarkable capabilities in a wide range of vision-language tasks, such as visual question answering (VQA) and complex reasoning\cite{vqasurvey,emoverse,zero-shot}.
In these models, an image is processed by a vision encoder\cite{clip,siglip} and a projector\cite{blip2,honeybee} to generate a sequence of visual tokens, which are then fed into the LLM\cite{llama,qwen,internlm2}.
However, this introduces significant computational overhead.
In a typical input, the number of visual tokens can reach hundreds, often far exceeding the length of text tokens.
Since the computational complexity of the self-attention mechanism scales quadratically with the sequence length, a large number of visual tokens makes using MLLMs costly.
When processing high-resolution images\cite{llava-next,internvl1.5,llava-onevision} or video streams\cite{video-llava,video-xl-pro,internvideo2.5}, the number of tokens increases even further.
Therefore, developing effective visual token pruning strategies is crucial for enhancing the inference efficiency of MLLMs.

\begin{figure}[t]
    \centering
    \includegraphics[width=\columnwidth]{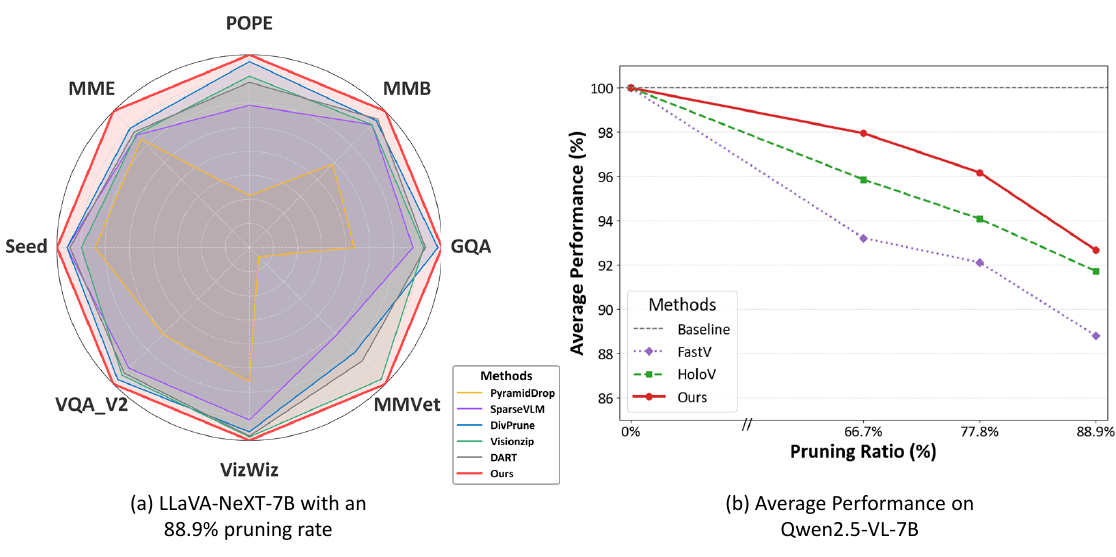}
    \caption{Performance comparison of GridPrune against state-of-the-art methods across various MLLM architectures.
    (a) presents results on the high-resolution LLaVA-NeXT-7B, with 11.1\% of visual tokens retained.
    (b) shows the average performance trend on Qwen2.5-VL-7B as the token retention ratio varies.}
    \label{performance}
\end{figure}

\begin{figure*}[t!]
    \centering
    \includegraphics[width=\textwidth]{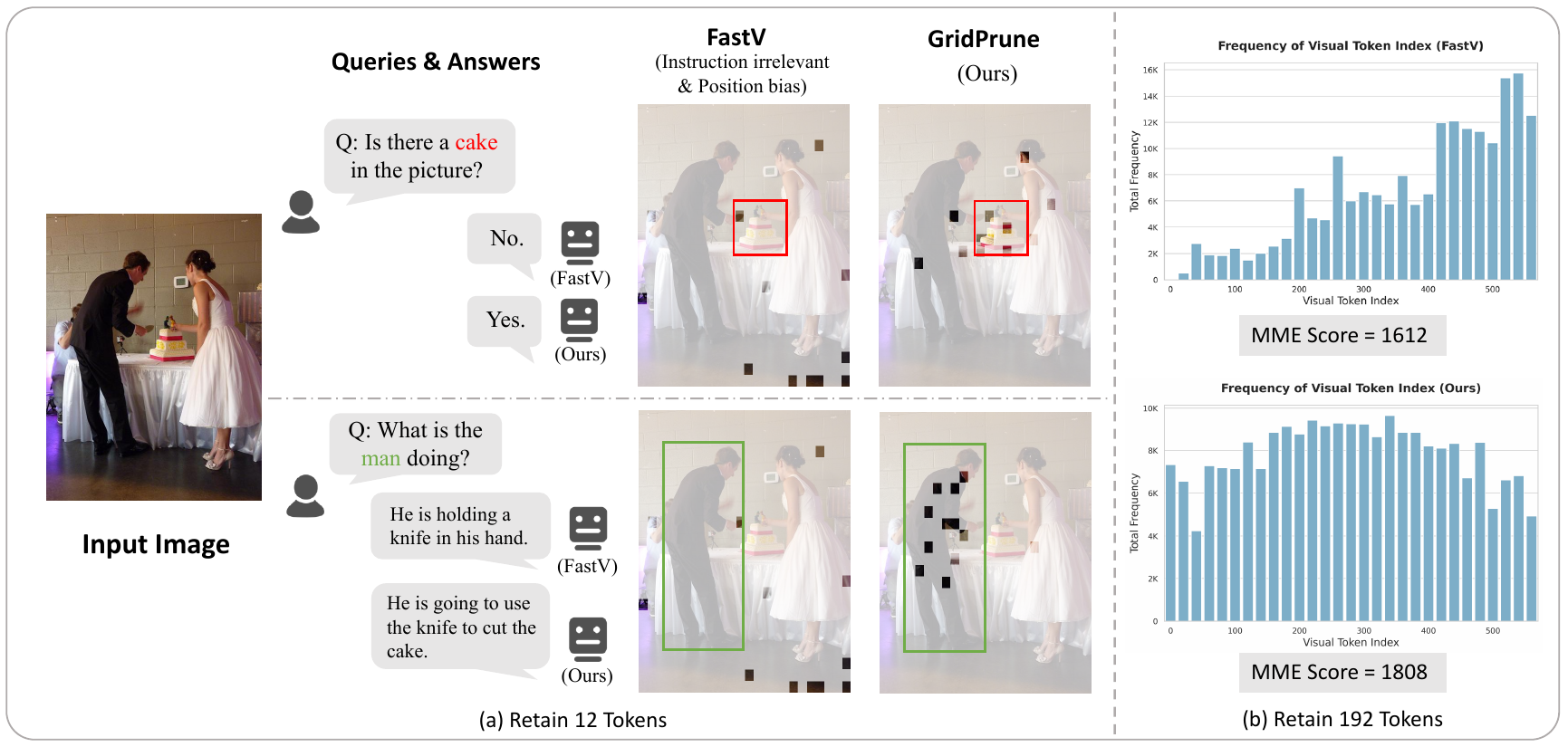}
    \caption{Comparison of GridPrune with FastV.
    (a) In a direct comparison, FastV’s selection is guided by a positional bias towards final tokens, while GridPrune’s is guided by the query’s semantic content.
    (b) Statistical analysis on the MME benchmark at scale shows that the histogram of selected indices exhibits a massive spike for FastV at the end of the sequence, revealing a strong positional bias.
    This is inefficient, as important content in images is typically centered or evenly distributed, rather than confined to one corner.
    In contrast, GridPrune's distribution is more balanced.}
    \label{compare}
\end{figure*}

Extensive efforts have been made in visual token pruning to reduce the inference cost of MLLMs\cite{transprune,fastv,sparsevlm}.
Early mainstream methods can be broadly categorized into two types: attention-based\cite{fastv,pyramiddrop,sparsevlm} and similarity-based\cite{tome,dart,divprune}.
Attention-based methods typically rely on attention scores as an importance signal for token selection.
However, they suffer from two issues: first, they are susceptible to position bias in global ranking, leading to unreliable selection results\cite{acl_pruning,holov,cdpruner}, as shown in Fig. \ref{compare} (a); 
second, the selected tokens are often highly redundant\cite{holov}, because nearby tokens with similar visual features tend to receive same high attention scores.
On the other hand, similarity-based methods, which mainly aim to construct a generic token subset, are often instruction-irrelevant, resulting in low informational efficiency when handling specific tasks\cite{cdpruner,visionzip}.
Subsequent work attempted to improve this from different angles.
However, these methods either pursue diversity through task conditioning at the global level\cite{cdpruner}, which lacks effective spatial planning, or rely on crops that are task-unaware and cannot be dynamically adjusted to specific questions\cite{holov}.
\textbf{Overall, existing research has focused mainly on how to optimize the ``what to select" problem, but has overlooked another problem, ``where to look"}, which can lead to inefficient spatial allocation, positional bias, and the retention of irrelevant or redundant tokens.

Studies in cognitive science provide inspiration for this challenge.
Research shows that when observing a scene, humans tend to first determine ``where to look" before deciding ``what to select" \cite{theory}.
This two-stage strategy allows the visual system to efficiently allocate attention at a coarse spatial level before performing fine-grained selection.
We hypothesize that the problems in current methods, like positional bias and redundancy, are caused by neglecting the ``where to look" step.
This is because deciding ``where to look" first allows the model to focus its budget on important regions, instead of wasting it on irrelevant backgrounds.
This makes the ``what to select" step more effective.

In this paper, we propose GridPrune, a pruning method that incorporates this two-stage strategy into MLLMs.
GridPrune splits the pruning process into two steps.
First, it uses the text query as a high-level command to make a task-driven decision, dynamically giving the limited token budget to image zones that are relevant to the task, which solves the ``where to look" problem.
Second, inside each zone that gets a budget, a selection process is carried out to pick the most informative tokens based on a fused score of text relevance and visual saliency, which solves the ``what to select" problem.
GridPrune changes a global optimization problem into a set of local decisions, which not only reduces problems like position bias but also improves information efficiency by giving the token budget to instruction-relevant locations, as shown in Fig. \ref{compare}.
Additionally, our method adapts to high-resolution application scenarios.
We test the effectiveness of GridPrune on ten benchmarks, demonstrating that GridPrune outperforms existing state-of-the-art approaches, especially under aggressive pruning ratios, as shown in Fig. \ref{performance}.
For instance, on LLaVA-NeXT-7B (Table \ref{tab:LLaVA-NeXT-7B}), GridPrune retains 96.98\% of the original performance while using only 11.1\% of the tokens, which is 2.34\% higher than the best-performing baseline. 
At a lower retention rate of 5.6\%, GridPrune outperforms the best-performing baseline by 3.1\% in average performance.

In summary, our contributions are summarized as follows:
\begin{itemize}
\item We identify the limitation of existing visual token pruning methods: they primarily focus on the ``what to select" problem, while rarely considering ``where to look", which leads to issues such as inefficient spatial allocation and positional bias.
\item We propose GridPrune, a training-free method based on above two-stage method. 
It divides the pruning process into a task-driven budget allocation and a series of intra-zone selections.
\item Extensive experiments show that GridPrune performs better than state-of-the-art methods across diverse MLLM architectures.
It can keep most of the model's original performance even with very few tokens.
\end{itemize}

\begin{figure*}[htbp]
    \centering
    \includegraphics[width=\textwidth]{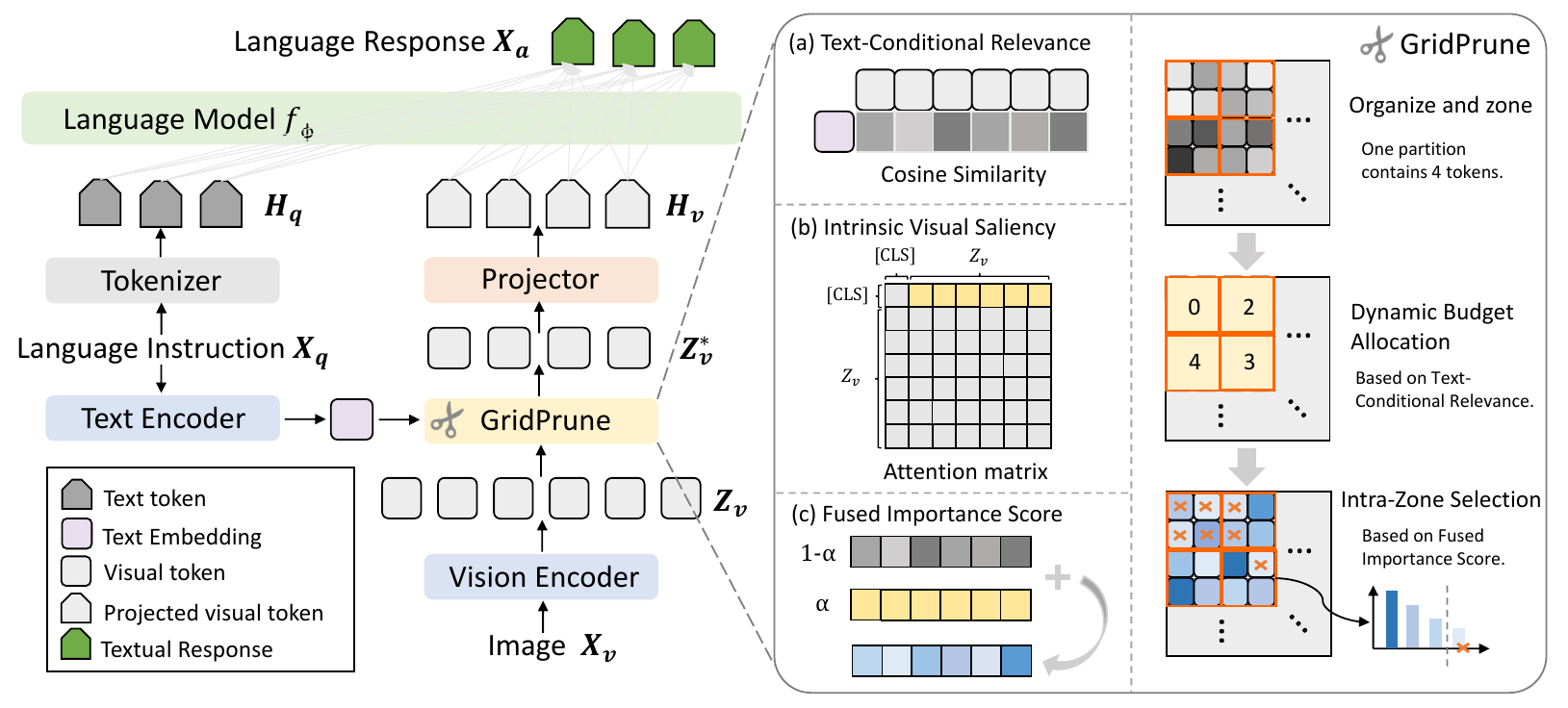}
    \caption{An overview of the GridPrune method.
    We first calculate two scores for each visual token: (a) Text-Conditional Relevance, derived from the cosine similarity between token embeddings and the text embedding (obtained from the CLIP text encoder using the user's prompt as input), and (b) Intrinsic Visual Saliency, extracted from the vision encoder's attention matrix.
    These are combined into (c) Fused Importance Score via \(\alpha\).
    GridPrune follows a ``guide-globally, select-locally" process: (1) the tokens are partitioned into zones; (2) a token budget is dynamically allocated to these zones based on their aggregate text-conditional relevance; and (3) a local Top-K selection is performed within each zone using the fused importance score to select the final token set.
    This mechanism ensures a selection that is both query-aware and spatially balanced.}
    \label{method}
\end{figure*}

\section{Related Work}

\subsection{Multimodal large language models.}
The remarkable achievements of Large Language Models (LLMs)\cite{llama,qwen,internlm2} have encouraged the development of Multimodal Large Language Models (MLLMs)\cite{llava,llava1.5,qwen2.5vl,internvl,instructblip,minigpt4}.
The dominant paradigm for these models is to connect a pretrained vision encoder\cite{clip,siglip} to an LLM via a projector\cite{blip2,honeybee}.
This architecture converts visual inputs into a token sequence to take advantage of the powerful capabilities of the LLM.
However, this introduces significant computational challenges.
The token sequence encoded from the images is typically much longer than the text input\cite{fastv,pyramiddrop}.
For example, LLaVA-1.5\cite{llava1.5} converts a standard image into 576 visual tokens.
The number of tokens increases further in scenarios that involve higher-resolution images\cite{llava-next,internvl1.5,llava-onevision} or video streams\cite{video-llava,video-xl-pro,internvideo2.5}.
These lengthy visual tokens lead to high inference costs, because the self-attention mechanism in LLMs has a computational complexity that scales quadratically with sequence length.
Therefore, developing efficient visual token pruning strategies is crucial for improving the inference efficiency of MLLMs.

\subsection{Token Pruning.}
To reduce the computational burden of MLLMs, researchers have made many efforts in visual token pruning \cite{transprune,fastv,sparsevlm}.
Early mainstream methods can be roughly divided into two types: attention-based \cite{fastv,pyramiddrop,sparsevlm} and similarity-based \cite{tome,dart,divprune}.

Attention-based methods rely on attention scores as the main importance signal.
FastV \cite{fastv} computes the average attention score that a token receives from all other tokens to judge its importance;
PyramidDrop \cite{pyramiddrop} uses a multi-stage pruning strategy;
and SparseVLM \cite{sparsevlm} uses the attention between instruction tokens and visual tokens.
However, this type of method has two main problems:
first, it is easily affected by position bias in global ranking, which leads to unreliable selection results \cite{acl_pruning,holov,cdpruner};
second, the selected tokens are often highly redundant, because nearby tokens tend to receive similarly high scores \cite{holov}.

Similarity-based methods focus on building a general token subset.
ToMe \cite{tome} progressively combines similar tokens through token merging;
DART \cite{dart} aims to select a representative subset from the feature space;
and DivPrune \cite{divprune} frames the problem as a Max-Min Diversity Problem to maximize the diversity among selected tokens.
VisionZip \cite{visionzip} also merges similar tokens to reduce redundancy.
The main limitation of these methods is that they are often instruction-irrelevant, which leads to low efficiency when processing task-specific queries \cite{cdpruner,visionzip}.

Later studies attempted to improve this from different perspectives.
CDPruner \cite{cdpruner} introduced text conditioning to guide diversity selection, but the text only serves as a way to adjust for diversity.
This limits the potential for the task instruction to perform spatial planning.
HoloV \cite{holov} uses spatial crops to preserve context, but its strategy is task-unaware.
Overall, existing research has mainly focused on how to score or diversify individual tokens (``what to select").
In contrast, how to allocate the limited token budget spatially (``where to look") has received less attention, which can lead to inefficient spatial allocation, positional bias, and the retention of irrelevant or redundant tokens.

\section{Method}
\label{sec:method}

This section introduces the proposed GridPrune method.
We first redefine the visual token pruning problem, and then introduce the two core components of GridPrune: a Dual-Source Importance Scoring function and the ``guide-globally, select-locally" zonal selection system, as shown in Fig. \ref{method}.

\subsection{Preliminaries and Problem Formulation}
\label{sec:3.1}
In MLLMs, a vision encoder transforms an image into a large set of $N$ patch tokens, $V = \{v_1, \dots, v_N\}$.
The goal of visual token pruning is to select a small subset $V' \subset V$ of size $k \ll N$ to reduce the computational load, which is dominated by the $O(N^2)$ self-attention complexity in the subsequent LLM.

The current pruning paradigm operates on the principle of global importance.
It first assigns a scalar importance score $s_i$ to each token $v_i$ and then performs a global Top-K selection:

\begin{equation}
V' = \underset{V_{sub} \subset V, |V_{sub}|=k}{\text{argmax}} \sum_{v_i \in V_{sub}} s_i
\end{equation}

\begin{algorithm}[t]
    \caption{Budget Allocation}
    \label{alg:budget_allocation}
    \begin{algorithmic}[1]
        \Require Vector of float budgets $B=\{b_1, \dots, b_M\}$; retained tokens $k$; zone capacity $c=block\_size^2$.
        \Ensure Vector of integer budgets $K=\{k_1, \dots, k_M\}$.
        \Statex

        \State $K_j \gets \min(\lfloor B_j \rfloor, c), \quad \forall j \in \{1, \dots, M\}$
        \Comment{Initialize integer budgets with capped floor values}
        
        \State $k' \gets k - \sum_{j=1}^{M} K_j$
        \Comment{Calculate the number of unallocated tokens}

        \State $F_j \gets B_j - \lfloor B_j \rfloor, \quad \forall j \in \{1, \dots, M\}$
        \Statex

        \For{$i = 1 \to k'$}
            \State $j^* \gets \underset{j \text{ s.t. } K_j < c}{\operatorname{argmax}} \, F_j$
            \Comment{Find the zone with the highest priority}

            \State $K_{j^*} \gets K_{j^*} + 1$
            \Comment{Allocate one token to the selected zone}

            \State $F_{j^*} \gets -\infty$
            \Comment{Prevent re-selection of the same zone}
        \EndFor
        \Statex
        \State \Return $K$
    \end{algorithmic}
\end{algorithm}

This global sorting method, although intuitive, has its limitations.
For example, attention-based methods often derive the scores $s_i$ from ViT\cite{vit}, which are affected by positional bias, leading the selection process to systematically over-sample from certain spatial regions while neglecting others.
Instead of seeking the globally ``most important" tokens, we propose a two-stage process: first, use text-conditional guidance to dynamically allocate a token budget across spatial zones, and then perform local selection within each budgeted zone.

\subsection{Dual-Source Importance Score}
\label{sec:3.2}
In order to achieve effective selection within our two-stage system, we first require a robust scoring function.
Therefore, we propose a Dual-Source Importance Scoring function that combines two complementary sources of information: the conditional relevance to the user's query and the model's intrinsic visual saliency.

\textbf{Text-Conditional Relevance.}
To make the pruning process respond to the user's prompt, we compute a relevance score $r_i$ for each patch token.
We leverage the hidden states $\{h_i\}$ from CLIP's penultimate layer, as they are already enriched with contextual information.
These hidden states are projected using the vision tower's own projection layer, $g_v$, to obtain multimodal visual embeddings.
The relevance score $r_i$ is then the cosine similarity between this projected embedding and the text embedding (the [EOS] token \cite{less_is_more,emu3}) from the CLIP text encoder $f_t$:

\begin{equation}
\label{eq:relevance_score}
r_i = \frac{g_v(h_i) \cdot f_t(Q)}{\|g_v(h_i)\| \cdot \|f_t(Q)\|}
\end{equation}

\begin{algorithm}[t!]
    \caption{GridPrune for a High-Resolution Image}
    \label{alg:high-resolution}
    \begin{algorithmic}[1]
        \Require A high-resolution image $I$, processed into a set of $S$ sub-images (e.g., 4 high-res tiles + 1 low-res global view), $I = \{p_1, \dots, p_S\}$. A single text prompt $T$.
        \Ensure The final visual feature sequence $F'$.
        \Statex

        \ForAll{$i \in \{1, \dots, S\}$}
            \State $p'_i \gets \text{GridPrune}(p_i, T)$
            \Comment{Apply pruning to each sub-image}
        \EndFor
        \Statex

        \State $P'' \gets \{p'_1, \dots, p'_S\}$
        \Comment{Collect pruned token sets}
        
        \State $\tilde{P} \gets g_{\text{mm\_proj}}(P'')$
        \Comment{Project all features in one batch}

        \State $F' \gets \bigoplus_{i=1}^{S} \tilde{p}_i$
        \Comment{Concatenate features into final sequence}
        \Statex
        
        \State \Return $F'$
    \end{algorithmic}
\end{algorithm}

This operation is carried out completely independently of the visual tower module. 
By using a highly text-aware representation method, we ensure that the relevance score depends on the user's intention.

\textbf{Intrinsic Visual Saliency.}
While text relevance is important, the model's intrinsic understanding of salient regions provides another valuable signal.
We derive this signal from the multi-head self-attention mechanism in the penultimate layer of the CLIP vision encoder.
Specifically, we obtain the attention weights from the [CLS] token to all patch tokens, and average these weights across all attention heads.
The scores are then min–max normalized to the $[0, 1]$ range.
The final score $a_i$ reflects the model's assessment of which vision token holds the most information for the overall image representation.

\textbf{Fused Importance Score.}
Finally, we fuse these two scores into a single importance score $s_i$.
To ensure both scores are on a comparable scale, we first normalize the text-conditional relevance score $r_i$ from its native range of $[-1, 1]$ to $[0, 1]$:

\begin{equation}
\hat{r_i} = (r_i + 1) / 2
\end{equation}

The final fused score $s_i$ is then a weighted linear combination of the normalized relevance score $\hat{r_i}$ and the saliency score $a_i$:

\begin{equation}
s_i = (1 - \alpha) \cdot \hat{r_i} + \alpha \cdot a_i
\end{equation}

where the hyperparameter $\alpha \in [0, 1]$ balances the influence of text-conditional guidance and intrinsic visual saliency.
As suggested by prior work \cite{acl_pruning}, the relevance and saliency score scales are not consistent across different architectures, and the ideal trade-off shifts with the pruning rate.
Therefore, we tune this parameter for each model and rate, which leads to a more effective balance and better overall performance.

These scores lay the foundation for our two-stage selection strategy.
The text-conditional relevance ($r_i$) is used in the ``where to look" stage, while the fused importance score ($s_i$) is used in the ``what to select" stage.
This design mimics how the brain operates: $r_i$ first provides a task-driven answer for ``where to look", preventing irrelevant but visually prominent areas from competing for the budget.
Then, $s_i$ solves the problem of ``what to select" by integrating all the information in these areas.

\begin{figure}[t]
    \centering
    \includegraphics[width=\columnwidth]{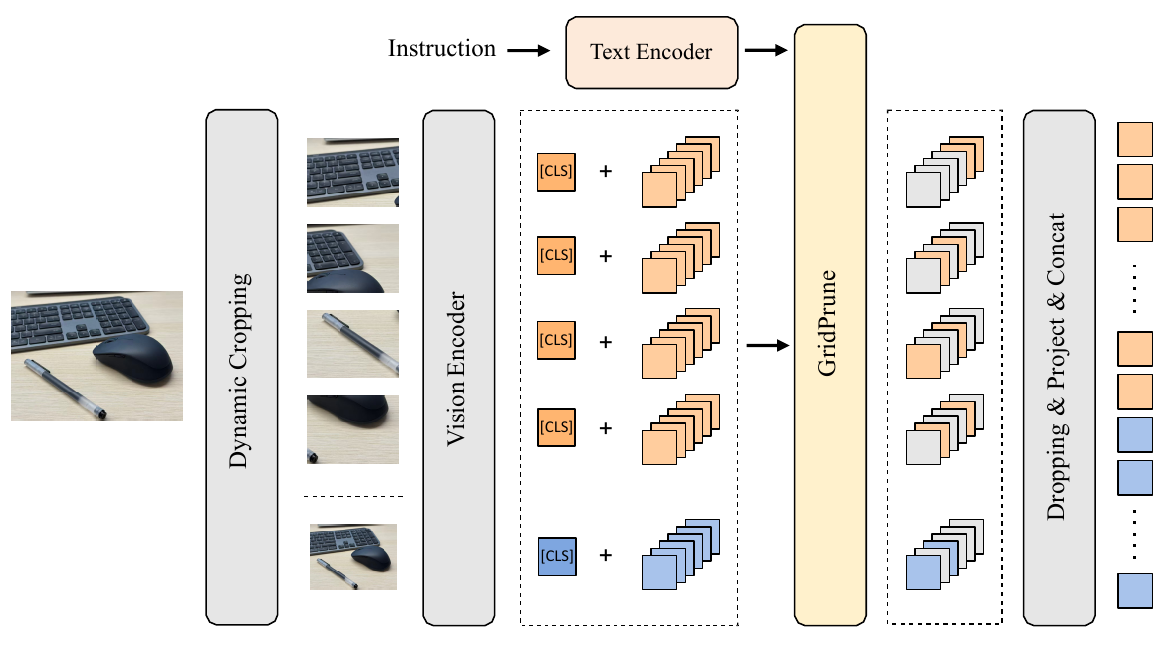}
    \caption{The processing flow of GridPrune applied to LLaVA-NeXT.
    The image is first dynamically cropped into multiple sub-images, and each sub-image independently passes through the vision encoder.
    Then, guided by the instruction, GridPrune prunes the tokens of each sub-image separately.
    Finally, all the retained tokens are projected and concatenated to form the final sequence.}
    \label{high}
\end{figure}

\subsection{Zonal Selection System}
\label{sec:3.3}
The proposed ``guide-globally, select-locally" zonal selection system replaces the conventional global Top-K mechanism, enabling a more spatially aware and efficient allocation of the token budget. 
First, it uses text-conditional guidance to dynamically allocate a token budget across spatial zones, and then performs local selection within each budgeted zone.

\textbf{Zone Partitioning.}
The image's $N$ patch tokens are organized into a 2D grid and partitioned into $M$ non-overlapping square zones of a predefined block size.
To determine the budget for each zone, we compute an aggregate relevance score, $\bar{r}_j$, by averaging the text-conditional relevance scores $r_i$ of all tokens within zone $j$.

\textbf{Dynamic Budget Allocation.}
We dynamically allocate the total token budget $k$ across the $M$ zones.
This is accomplished by converting the zone scores $\{\bar{r}_j\}$ into a probability distribution using a standard softmax function:

\begin{equation}
\label{eq:softmax}
P_j = \frac{\exp(\bar{r}_j)}{\sum_{m=1}^{M} \exp(\bar{r}_m)}
\end{equation}

The budget $k_j$ for each zone is then determined by a rounding procedure applied to $P_j \cdot k$ that ensures $\sum k_j = k$, detailed in Algorithm \ref{alg:budget_allocation}.
This step translates text relevance into a spatial allocation plan.

\textbf{Intra-Zone Selection with Fused Score.}
With the budget $k_j$ for each zone finalized, the final selection is made locally.
Within each zone $j$, we select the $k_j$ tokens that have the highest fused importance scores $s_i$.
By separating the query-aware `where to look' step from the local `what to select' step, this two-stage process addresses the limitations of global Top-K selection, such as inefficient spatial allocation, positional bias, and redundancy.

\begin{table}[h!]
  \centering
  \small
  \renewcommand{\arraystretch}{0.85}
  \caption{Optimal settings for the fusion hyperparameter \(\alpha\) used across all main experiments.}
  \label{tab:alpha_settings}
  \begin{tabular}{l | c c}
    \toprule
    \textbf{Base Model} & \textbf{Retention ratio} & \textbf{Value of $\alpha$} \\
    \midrule
    \rowcolor{lightgray}
    \multicolumn{3}{c}{\textit{Standard Resolution}} \\
    \multirow{3}{*}{LLaVA-1.5-7B} & 33.3\% & 0.8 \\
                                  & 22.2\% & 0.7 \\
                                  & 11.1\%  & 0.7 \\
    \midrule
    \rowcolor{lightgray}
    \multicolumn{3}{c}{\textit{High Resolution}} \\
    \multirow{3}{*}{LLaVA-NeXT-7B} & 22.2\% & 0.8 \\
                                   & 11.1\% & 0.7 \\
                                   & 5.6\% & 0.7 \\
    \midrule
    \rowcolor{lightgray}
    \multicolumn{3}{c}{\textit{Qwen Architecture}} \\
    \multirow{3}{*}{Qwen2.5-VL-7B} & 33.3\% & 0.7 \\
                                   & 22.2\% & 0.7 \\
                                   & 11.1\% & 0.7 \\
    \bottomrule
  \end{tabular}
\end{table}

\section{Experiments}

\subsection{Experimental Setup}

\textbf{Models and Baselines.}
To validate the effectiveness and generalizability of GridPrune, we conduct experiments on a diverse set of three multimodal large language models.
LLaVA-1.5-7B\cite{llava1.5} serves as a widely adopted benchmark model for standard-resolution inputs.
LLaVA-NeXT-7B\cite{llava-next} represents models designed for high-resolution imagery, processing a larger sequence of tokens.
Qwen2.5-VL-7B\cite{qwen2.5vl}, featuring a distinct architecture from the LLaVA series, is included to assess the architectural generalization of our method.
We compare GridPrune against a comprehensive set of state-of-the-art (SOTA) training-free pruning methods, including ToMe\cite{tome}, FastV\cite{fastv}, PyramidDrop\cite{pyramiddrop}, SparseVLM\cite{sparsevlm}, DivPrune\cite{divprune}, VisionZip\cite{visionzip}, LLaVA-PruMerge\cite{llava-prumerge}, DART\cite{dart}, and HoloV\cite{holov}.

\begin{table}[htbp]
  \centering
  \caption{Performance comparison on Qwen2.5-VL-7B. Avg.(\%) represents the average percentage of performance maintained.}
  \label{tab:Qwen2.5-VL-7B}
  \resizebox{\columnwidth}{!}{%
    \begin{tabular}{l c | c c c c}
      \toprule
      \textbf{Methods} & \textbf{Avg.(\%)} & \textbf{SQA\textsuperscript{IMG}} &  \textbf{POPE} & \textbf{MMB\textsuperscript{EN}} & \textbf{MME} \\
      \midrule
      \rowcolor{lightgray}
      \multicolumn{6}{c}{\textbf{\textit{Upper Bound (100\%)}}} \\
      Qwen2.5-VL-7B & 100\% & 84.7 & 86.1 & 82.8 & 2304 \\
      \midrule
      \rowcolor{lightgray}
      \multicolumn{6}{c}{\textbf{\textit{Retain 33.3\% Tokens}}} \\
      FastV (ECCV24) & 92.4\% & 78.5 & 82.2 & 75.7 & 2072 \\
      HoloV (NeurIPS25) & 94.6\% & 79.8 & 85.0 & 78.3 & 2093 \\
      \rowcolor{green!15}
      \textbf{GridPrune (Ours)} & \textbf{97.6\%} & 84.4 & 84.2 & 79.8 & 2221 \\
      \midrule
      \rowcolor{lightgray}
      \multicolumn{6}{c}{\textbf{\textit{Retain 22.2\% Tokens}}} \\
      FastV (ECCV24) & 91.2\% & 78.0 & 80.7 & 74.9 & 2036 \\
      HoloV (NeurIPS25) & 92.7\% & 79.8 & 82.3 & 76.5 & 2043 \\
      \rowcolor{green!15}
      \textbf{GridPrune (Ours)} & \textbf{95.6\%} & 83.1 & 82.8 & 78.0 & 2161 \\
      \midrule
      \rowcolor{lightgray}
      \multicolumn{6}{c}{\textbf{\textit{Retain 11.1\% Tokens}}} \\
      FastV (ECCV24) & 87.6\% & 77.4 & 78.6 & 69.2 & 1940 \\
      HoloV (NeurIPS25) & 90.5\% & 79.5 & 80.7 & 72.4 & 2006 \\
      \rowcolor{green!15}
      \textbf{GridPrune (Ours)} & \textbf{91.3\%} & 80.7 & 78.1 & 76.2 & 2009 \\
      \bottomrule
    \end{tabular}
  }
\end{table}

\begin{table*}[t!]
  \centering
  \caption{Main results on the standard-resolution benchmark LLaVA-1.5-7B.}
  \label{tab:LLaVA-1.5-7B}
  \setlength{\tabcolsep}{5pt}
  \resizebox{\textwidth}{!}{%
  \begin{tabular}{l c | c c c c c c c c c c}
    \toprule
    \textbf{Methods} & \textbf{Avg.(\%)} & \textbf{SQA\textsuperscript{IMG}} & \textbf{GQA} & \textbf{MMB\textsuperscript{EN}} & \textbf{POPE} & \textbf{TextVQA} & \textbf{MME} & \textbf{Seed\textsuperscript{I}} & \textbf{VQA\textsuperscript{V2}} & \textbf{VizWiz} & \textbf{MMVet} \\
    \midrule
    \rowcolor{lightgray}
    \multicolumn{12}{c}{\textbf{\textit{Upper Bound, 576 Tokens (100\%)}}} \\
    LLaVA-1.5-7B & 100\% & 69.5 & 61.9 & 64.6 & 85.9 & 58.2 & 1862 & 66.2 & 78.5 & 50.1 & 31.1 \\
    \midrule
    \rowcolor{lightgray}
    \multicolumn{12}{c}{\textbf{\textit{Retain 192 Tokens (33.3\%)}}} \\
    ToMe (ICLR23) & - & 65.2 & 54.3 & 60.5 & 72.4 & 52.1 & 1563 & - & 68.0 & - & - \\
    FastV (ECCV24) & 89.13\% & 67.3 & 52.7 & 61.2 & 64.8 & 52.5 & 1612 & 57.2 & 67.1 & 50.8 & 27.7 \\
    PyramidDrop (CVPR25) & 96.20\% & 69.2 & 57.1 & 63.2 & 82.3 & 56.1 & 1797 & 58.2 & 75.1 & 51.1 & 30.5 \\
    DivPrune (CVPR25) & 98.95\% & 68.9 & 59.9 & 62.3 & 87.0 & 56.9 & 1762 & 64.2 & 76.8 & 54.9 & 30.8 \\
    Visionzip (CVPR25) & 98.30\% & 68.8 & 59.3 & 62.9 & 85.5 & 57.2 & 1769 & 63.2 & 76.8 & 51.5 & 31.7 \\
    SparseVLM (ICML25) & 97.45\% & 69.1 & 57.6 & 62.5 & 83.6 & 56.1 & 1787 & 64.2 & 75.6 & 50.6 & 31.5 \\
    LLaVA-PruMerge (ICCV25) & - & 67.9 & 54.3 & 59.6 & 71.3 & 54.3 & 1632 & - & 70.6 & 50.1 & - \\
    DART (EMNLP25) & 98.61\% & 69.8 & 58.9 & 63.6 & 82.8 & 57.4 & 1834 & 64.6 & 76.7 & 51.1 & 31.5 \\
    HoloV (NeurIPS25) & - & 69.8 & 59.0 & 65.4 & 85.6 & 57.4 & 1820 & - & 76.7 & 50.9 & - \\
    \rowcolor{green!15}
    \textbf{GridPrune (Ours)} & \textbf{99.81\%} & 68.6 & 60.7 & 63.7 & 86.3 & 57.0 & 1808 & 64.5 & 77.4 & 52.1 & 33.3 \\
    \midrule
    \rowcolor{lightgray}
    \multicolumn{12}{c}{\textbf{\textit{Retain 128 Tokens (22.2\%)}}} \\
    ToMe (ICLR23) & - & 59.6 & 52.4 & 53.3 & 62.8 & 49.1 & 1343 & - & 63.0 & - & - \\
    FastV (ECCV24) & 86.94\% & 68.5 & 54.0 & 56.1 & 68.2 & 56.4 & 1490 & 52.2 & 71.0 & 51.9 & 27.0 \\
    PyramidDrop (CVPR25) & 93.30\% & 68.4 & 57.1 & 62.3 & 77.5 & 56.7 & 1761 & 54.1 & 74.3 & 49.4 & 27.6 \\
    DivPrune (CVPR25) & 97.43\% & 68.6 & 59.4 & 61.5 & 87.0 & 55.9 & 1718 & 62.4 & 76.0 & 52.8 & 30.6 \\
    Visionzip (CVPR25) & 97.47\% & 68.9 & 57.7 & 62.0 & 83.2 & 56.8 & 1757 & 61.3 & 75.6 & 52.0 & 32.6 \\
    SparseVLM (ICML25) & 96.24\% & 69.0 & 57.3 & 62.6 & 83.1 & 56.3 & 1746 & 63.6 & 75.1 & 49.7 & 29.7 \\
    LLaVA-PruMerge (ICCV25) & - & 67.1 & 53.3 & 58.1 & 67.2 & 54.3 & 1554 & - & 68.8 & 50.3 & 30.4 \\
    DART (EMNLP25) & 97.50\% & 69.1 & 57.9 & 63.2 & 80.1 & 56.4 & 1845 & 63.4 & 75.9 & 51.7 & 30.9 \\
    HoloV (NeurIPS25) & - & 69.8 & 57.7 & 63.9 & 84.0 & 56.8 & 1802 & - & 75.5 & 51.5 & - \\
    \rowcolor{green!15}
    \textbf{GridPrune (Ours)} & \textbf{98.12\%} & 68.5 & 59.6 & 62.4 & 86.2 & 54.9 & 1744 & 62.9 & 76.2 & 52.7 & 32.4 \\
    \midrule
    \rowcolor{lightgray}
    \multicolumn{12}{c}{\textbf{\textit{Retain 64 Tokens (11.1\%)}}} \\
    ToMe (ICLR23) & - & 50.0 & 48.6 & 43.7 & 52.5 & 45.3 & 1138 & - & 57.1 & - & - \\
    FastV (ECCV24) & 74.02\% & 68.4 & 46.0 & 50.1 & 35.5 & 51.6 & 1255 & 41.4 & 55.9 & 49.1 & 18.9 \\
    PyramidDrop (CVPR25) & 76.11\% & 69.0 & 46.1 & 48.0 & 40.8 & 50.6 & 1561 & 48.8 & 56.3 & 46.3 & 17.7 \\
    DivPrune (CVPR25) & 95.00\% & 68.0 & 57.5 & 60.1 & 85.5 & 54.5 & 1674 & 60.5 & 74.1 & 53.6 & 28.1 \\
    Visionzip (CVPR25) & 94.13\% & 69.0 & 55.1 & 60.1 & 77.0 & 55.5 & 1687 & 57.7 & 72.4 & 52.9 & 30.9 \\
    SparseVLM (ICML25) & 87.78\% & 69.2 & 52.0 & 58.3 & 69.7 & 52.1 & 1589 & 56.7 & 66.9 & 49.4 & 24.4 \\
    LLaVA-PruMerge (ICCV25) & - & 68.1 & 51.9 & 55.3 & 65.3 & 54.0 & 1549 & - & 67.4 & 50.1 & 28.0 \\
    DART (EMNLP25) & 92.88\% & 69.8 & 55.9 & 60.6 & 73.9 & 54.4 & 1765 & 59.3 & 72.4 & 51.6 & 26.5 \\
    HoloV (NeurIPS25) & - & 69.5 & 55.3 & 63.3 & 80.3 & 55.4 & 1715 & - & 72.8 & 52.8 & - \\
    \rowcolor{green!15}
    \textbf{GridPrune (Ours)} & \textbf{96.76\%} & 68.2 & 58.7 & 62.3 & 85.8 & 54.3 & 1719 & 62.1 & 75.3 & 54.5 & 29.3 \\
    \bottomrule
  \end{tabular}
  }
\end{table*}

\begin{table*}[t]
  \centering
  \caption{Performance on the high-resolution LLaVA-NeXT-7B model.}
  \label{tab:LLaVA-NeXT-7B}
  \setlength{\tabcolsep}{5pt}
    \resizebox{\textwidth}{!}{%
  \begin{tabular}{l c | c c c c c c c c c c}
    \toprule
    \textbf{Methods} & \textbf{Avg.(\%)} & \textbf{SQA\textsuperscript{IMG}} & \textbf{GQA} & \textbf{MMB\textsuperscript{EN}} & \textbf{POPE} & \textbf{TextVQA} & \textbf{MME} & \textbf{Seed\textsuperscript{I}} & \textbf{VQA\textsuperscript{V2}} & \textbf{VizWiz} & \textbf{MMVet} \\
    \midrule
    \rowcolor{lightgray}
    \multicolumn{12}{c}{\textbf{\textit{Upper Bound, 2880 Tokens (100\%)}}} \\
    LLaVA-NeXT-7B & 100\% & 70.2 & 64.3 & 67.9 & 86.5 & 61.3 & 1842 & 70.2 & 80.1 & 55.2 & 40.0 \\
    \midrule
    \rowcolor{lightgray}
    \multicolumn{12}{c}{\textbf{\textit{Retain 640 Tokens (22.2\%)}}} \\
    FastV (ECCV24) & 94.60\% & 67.4 & 58.9 & 63.1 & 79.5 & 58.1 & 1807 & 61.9 & 77.0 & 53.9 & 39.5 \\
    PyramidDrop (CVPR25) & 95.21\% & 66.7 & 60.0 & 64.1 & 83.8 & 57.8 & 1782 & 65.6 & 79.1 & 53.8 & 36.7 \\
    SparseVLM (ICML25) & 96.44\% & 67.6 & 61.2 & 65.9 & 85.3 & 59.7 & 1772 & 68.4 & 79.2 & 53.6 & 36.1 \\
    DivPrune (CVPR25) & 97.07\% & 67.8 & 61.9 & 65.8 & 86.9 & 57.0 & 1773 & 67.6 & 79.3 & 55.7 & 38.0 \\
    Visionzip (CVPR25) & 97.83\% & 68.1 & 61.3 & 66.3 & 86.2 & 59.9 & 1782 & 66.7 & 79.1 & 57.1 & 38.8 \\
    DART (EMNLP25) & 97.10\% & 68.2 & 61.3 & 64.9 & 85.0 & 59.5 & 1793 & 68.1 & 78.3 & 57.0 & 36.9 \\
    \rowcolor{green!15}
    \textbf{GridPrune (Ours)} & \textbf{98.30\%} & 68.2 & 62.8 & 67.1 & 86.9 & 57.4 & 1815 & 67.8 & 79.3 & 57.1 & 39.1 \\
    \midrule
    \rowcolor{lightgray}
    \multicolumn{12}{c}{\textbf{\textit{Retain 320 Tokens (11.1\%)}}} \\
    FastV (ECCV24) & 77.72\% & 66.6 & 49.8 & 53.4 & 49.5 & 52.2 & 1539 & 56.6 & 61.5 & 51.3 & 20.0 \\
    PyramidDrop (CVPR25) & 81.72\% & 66.7 & 50.4 & 55.5 & 60.8 & 49.0 & 1672 & 61.5 & 66.8 & 49.7 & 24.0 \\
    SparseVLM (ICML25) & 92.11\% & 67.2 & 57.9 & 63.1 & 76.9 & 56.5 & 1747 & 65.4 & 74.6 & 54.2 & 32.8 \\
    DivPrune (CVPR25) & 94.64\% & 67.7 & 61.1 & 63.9 & 84.7 & 56.2 & 1731 & 65.4 & 77.2 & 55.6 & 34.8 \\
    Visionzip (CVPR25) & 94.60\% & 67.3 & 59.3 & 63.1 & 82.1 & 58.9 & 1698 & 63.4 & 76.2 & 56.2 & 37.8 \\
    DART (EMNLP25) & 94.17\% & 67.5 & 59.5 & 64.2 & 81.0 & 57.6 & 1710 & 65.0 & 75.7 & 56.1 & 35.7 \\
    \rowcolor{green!15}
    \textbf{GridPrune (Ours)} & \textbf{96.98\%} & 67.3 & 61.6 & 65.7 & 85.9 & 56.5 & 1821 & 66.8 & 78.2 & 56.6 & 38.3 \\
    \midrule
    \rowcolor{lightgray}
    \multicolumn{12}{c}{\textbf{\textit{Retain 160 Tokens (5.6\%)}}} \\
    DivPrune (CVPR25) & 91.55\% & 67.1 & 59.3 & 62.9 & 80.0 & 54.1 & 1658 & 62.5 & 75.0 & 56.1 & 32.0 \\
    Visionzip (CVPR25) & 89.22\% & 68.3 & 55.5 & 60.1 & 74.8 & 55.7 & 1630 & 58.3 & 71.4 & 55.5 & 32.6 \\
    DART (EMNLP25) & 90.30\% & 67.8 & 56.8 & 62.0 & 75.3 & 54.9 & 1700 & 59.1 & 72.5 & 56.7 & 32.2 \\
    \rowcolor{green!15}
    \textbf{GridPrune (Ours)} & \textbf{94.65\%} & 66.7 & 60.0 & 64.2 & 85.5 & 54.0 & 1761 & 65.2 & 76.1 & 55.6 & 37.0 \\
    \bottomrule
  \end{tabular}
  }
\end{table*}

\textbf{Benchmarks.}
We conduct a comprehensive evaluation across a suite of ten challenging benchmarks to assess a wide range of multimodal capabilities.
The suite includes general visual question answering (VQAv2\cite{vqav2}, GQA\cite{gqa}, VizWiz\cite{vizwiz}); text-intensive VQA (TextVQA\cite{textvqa}); scientific reasoning (ScienceQA\cite{sqa}); compositional understanding (MME\cite{mme}, SEED-Bench\cite{seed}); object hallucination detection (POPE\cite{pope}); and holistic model assessment (MMBench\cite{mmbench}, MM-Vet\cite{mmvet}).

\textbf{Implementation Details.}
Across all experiments, GridPrune is configured with a block size of 2, partitioning the standard 24×24 patch grid into \( M = 144 \) zones.
The fusion hyperparameter \(\alpha\), which balances text-conditional guidance with intrinsic visual saliency, is tuned for different models and retention ratios, with specific values reported in Table \ref{tab:alpha_settings}.
The performance of the unpruned, full-token model is reported as the Upper Bound.
All methods are evaluated at multiple token retention ratios, and all experiments are conducted on RTX 3090 GPUs (24 GB).

\subsection{Main Results}

We present the main results of our comparative experiments in Tables \ref{tab:Qwen2.5-VL-7B}, \ref{tab:LLaVA-1.5-7B}, and \ref{tab:LLaVA-NeXT-7B}.
The findings across all models and benchmarks demonstrate the superior performance of GridPrune compared to existing SOTA methods.

\textbf{Performance on LLaVA-1.5-7B.}
As shown in Table \ref{tab:LLaVA-1.5-7B}, the average performance of GridPrune surpasses all other methods on the LLaVA-1.5-7B model.
At a retention ratio of 33.3\%, GridPrune achieves an average score of 99.81\% relative to the unpruned model, matching its performance while using only one-third of the visual tokens.
The advantage of GridPrune becomes even more pronounced under higher pruning settings.
When retaining 11.1\% of the tokens, GridPrune maintains a remarkable 96.76\% of the original performance.
In contrast, other methods experience a more evident performance drop, with GridPrune outperforming the best-performing baseline by 1.76 percentage points.
This demonstrates the robustness of our method in preserving critical visual information, even with a very small token budget.

\textbf{Performance on LLaVA-NeXT-7B.}
The algorithm \ref{alg:high-resolution} and Fig. \ref{high} show the logic of applying GridPrune to LLaVA-NeXT.
The effectiveness of GridPrune is evident in the high-resolution setting of LLaVA-NeXT-7B, where the initial number of tokens is much larger.
Table \ref{tab:LLaVA-NeXT-7B}  and Fig. \ref{performance} (a) show that the average performance of GridPrune outperforms all baselines.
For instance, when retaining 320 tokens, GridPrune achieves an average performance of 96.98\%, which is 2.34\% higher than that of DivPrune.
When only 160 tokens are kept, GridPrune still maintains 94.65\% of the upper-bound performance. 
This result shows that GridPrune enables high-resolution models to operate at a fraction of their original computational cost.

\textbf{Performance on Qwen2.5-VL-7B.}
To verify the architectural generalizability of our approach, we also evaluate GridPrune on Qwen2.5-VL-7B.
We calculate text relevance using the cosine similarity between the instruction vector (from the LLM's word embedding layer) and each visual token.
To obtain visual saliency, we extract the average self-attention score of each token from the last layer of the vision encoder.
The results are presented in Table \ref{tab:Qwen2.5-VL-7B} and Fig. \ref{performance} (b).
The average performance of GridPrune surpasses that of other methods across all tested retention ratios.
For example, at a 33.3\% retention rate, GridPrune achieves a 97.6\% average score, outperforming HoloV by 3 percentage points.
These results show that the principles behind GridPrune are not tailored to a specific model architecture but offer a more effective approach to token pruning.

\subsection{Efficiency Analysis.}

Our efficiency evaluation demonstrates the practical benefits of GridPrune, as shown in Table \ref{tab:efficiency}.
On LLaVA-1.5-7B, our method achieves a 2.14× speedup, reducing inference latency to 90.8 ms.
On LLaVA-NeXT-7B, GridPrune delivers a 5.09× speedup, cutting latency to just 113.4 ms.
These results show the substantial real-world acceleration provided by our method.

This empirical speedup is theoretically grounded in the reduction of total Floating-Point Operations (TFLOPs).
To quantify the computational cost associated with visual inputs, we follow the convention of prior works \cite{pyramiddrop,sparsevlm} and the total computational cost for the visual component across the $L$-layer decoder is approximated by:
\begin{equation}
\label{eq:flops_total_decoder}
\text{TFLOPs} \approx L \cdot (2N^2d + 4Nd^2 + 3Ndm)
\end{equation}
where $N$ is the number of visual tokens, $d$ is the hidden dimension, and $m$ is the FFN intermediate dimension.
By pruning visual tokens, GridPrune directly reduces $N$, primarily reducing the slowdown from the quadratic $2N^2d$ term.
This analysis of the visual processing cost formally justifies the empirical latency reduction observed in the full system, as the overhead of GridPrune is dwarfed by the computational savings in the decoder.

\begin{table}[htbp]
  \centering
  \small
  \caption{Efficiency and performance trade-off analysis.}
  \label{tab:efficiency}
  \setlength{\tabcolsep}{4pt}
  \begin{tabular}{l c | c c c}
    \toprule
    \textbf{Methods} & \textbf{Token↓} & \textbf{TFLOPs↓} & 
    \textbf{\begin{tabular}{@{}c@{}}Latency \\ (ms)↓\end{tabular}} & 
    \textbf{\begin{tabular}{@{}c@{}}MME \\ Score\end{tabular}}\\
    \midrule
    \rowcolor{lightgray}
    LLaVA-1.5-7B & 576 & 3.82 & 194.2 & 1862 \\
    + PyramidDrop & 192 & 1.30 & 112.4 & 1797 \\
    + SparseVLM & 192 & 1.33 & 100.5 & 1787 \\
    + Visionzip & 192 & \textbf{1.25} & 91.2 & 1769 \\
    \rowcolor{green!15}
    + GridPrune (Ours) & 192 & \textbf{1.25} & \textbf{90.8} & \textbf{1808} \\
    \midrule
    \rowcolor{lightgray}
    LLaVA-NeXT-7B & 2880 & 20.83 & 576.9 & 1842 \\
    + PyramidDrop & 320 & 3.02 & 227.3 & 1672 \\
    + SparseVLM & 320 & 3.04 & 150.9 & 1747 \\
    + Visionzip & 320 & \textbf{2.10} & 115.8 & 1698 \\
    \rowcolor{green!15}
    + GridPrune (Ours) & 320 & \textbf{2.10} & \textbf{113.4} & \textbf{1821} \\
    \bottomrule
  \end{tabular}
\end{table}

\begin{figure*}[t]
    \centering
    \includegraphics[width=\textwidth]{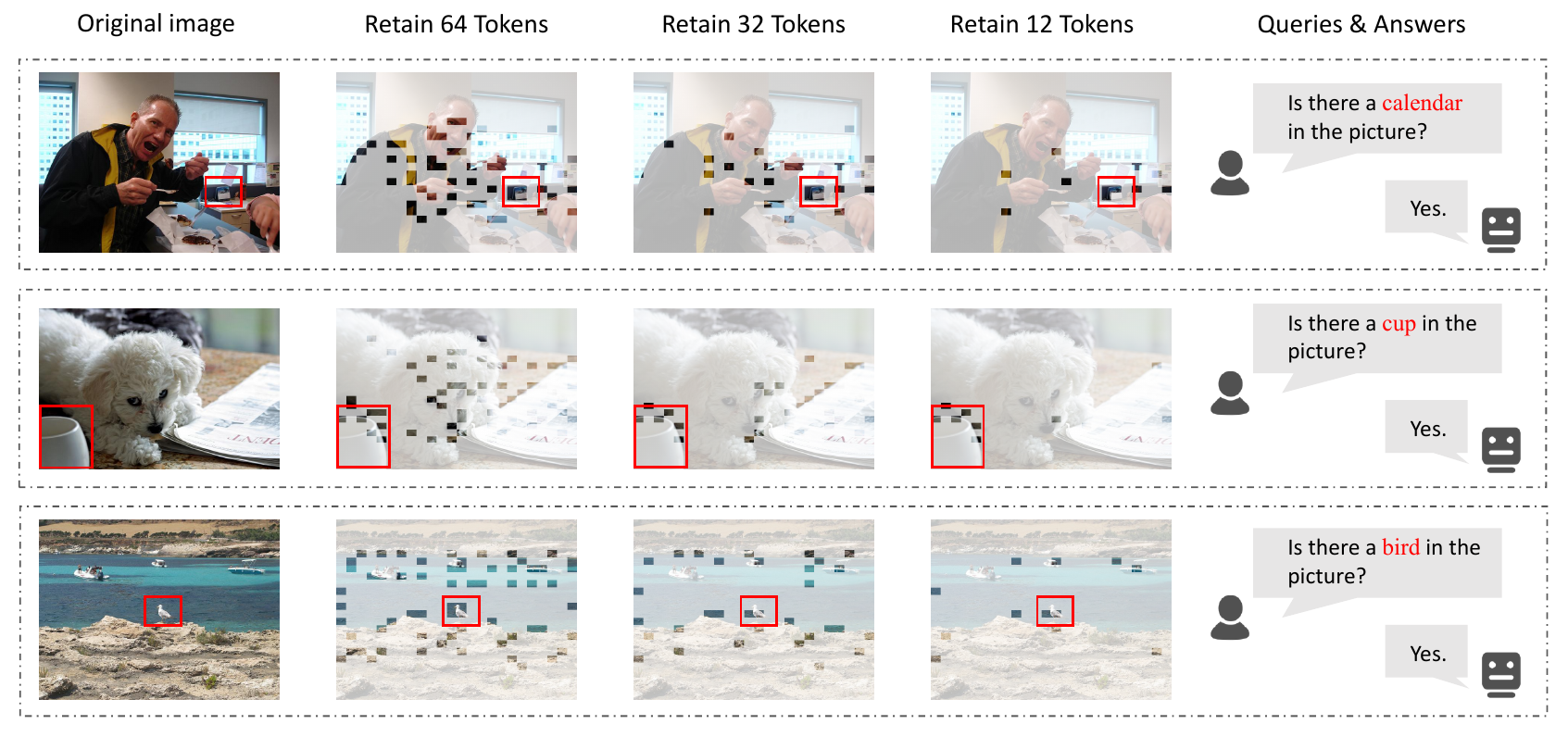}
    \caption{Visualization of GridPrune's token selection.
    The token focus dynamically shifts to the calendar, cup, or bird based on the user’s query, even when the number of reserved tokens is limited to 12.}
    \label{case}
\end{figure*}

\subsection{Ablation Studies}

To validate GridPrune, we conduct a series of ablation studies focusing on its two core components: the Zoned Selection Mechanism and the Fused Importance Score.

\textbf{Effectiveness of the Zoned Selection Mechanism.}
To verify this, we vary the block size to adjust the granularity of zone partitioning.
A larger block size results in fewer, coarser zones, while a smaller size creates more, finer-grained zones for local selection.
The results are presented in Table \ref{tab:block_size_ablation}.

The configuration with a block size of 24, which is equivalent to a standard global Top-K selection, serves as our baseline.
As the block size decreases, the overall performance improves.
Decreasing the block size from 24 to 2 leads to a steady increase in the average performance from 97.6\% to 98.6\%.
We argue that this is because finer partitioning provides a more granular budget allocation, which is crucial for preserving small but important details that might be overlooked in a coarser selection process.
While the optimal block size can exhibit minor variations across different benchmarks (e.g., size 3 slightly outperforms on MME), a block size of 2 achieves the highest average score and shows robust high performance across all tested benchmarks.
Considering its superior overall performance and stability, we adopt a block size of 2 as the default configuration for GridPrune.

\begin{table}[t]
  \centering
  \small
  \setlength{\tabcolsep}{3pt}
  \caption{Ablation study on the effect of selection granularity, conducted on LLaVA-1.5-7B with a fixed retention of 192 tokens and \(\alpha\) = 0.8.
  We vary the block size, where a size of 24 corresponds to global selection, and a size of 2 represents the most localized selection.}
  \label{tab:block_size_ablation}
    \resizebox{\columnwidth}{!}{%
  \begin{tabular}{c c c | c c c c c}
    \toprule
    \textbf{Size} & \textbf{Number} & \textbf{Avg.(\%)} & \textbf{MME} & \textbf{MMB\textsuperscript{EN}} & \textbf{POPE} & \textbf{SQA\textsuperscript{IMG}} & \textbf{GQA} \\
    \midrule
    24 & 1 & 97.6\% & 1767 & 62.5 & 86.6 & 68.7 & 59.8 \\
    12 & 4 & 97.6\% & 1767 & 63.3 & 86.5 & 68.1 & 59.8 \\
    8  & 9 & 97.9\% & 1777 & 63.2 & 86.6 & 68.5 & 60.1 \\
    4  & 36 & 98.3\% & 1788 & 63.6 & 86.1 & 69.0 & 60.3 \\
    3  & 64 & 98.4\% & 1809 & 63.9 & 86.0 & 68.2 & 60.6 \\
    \textbf{2}  & \textbf{144} & \textbf{98.6\%} & 1808 & 63.7 & 86.3 & 68.6 & 60.7 \\
    \bottomrule
  \end{tabular}
  }
\end{table}

\textbf{Analysis of the Fused Importance Score.}
GridPrune uses a fused score that combines text-conditional relevance and intrinsic visual saliency.
To analyze the contribution of each component, we ablate the fusion hyperparameter $\alpha$ from 0.0 to 1.0.
As shown in Table \ref{tab:alpha_ablation}, relying solely on text relevance or visual saliency does not produce the best results.
The optimal performance is achieved when $\alpha$ is set to 0.8 (for this specific model and retention rate), showing that both information sources are valuable.
This suggests that the model's own assessment of important visual features acts as a beneficial regularizer, preventing the model from overfitting to the text query and thus leading to a more robust and comprehensive final token set.

\begin{table}[t]
  \centering
  \small
  \setlength{\tabcolsep}{4pt} 
  \caption{Ablation study on the fusion hyperparameter \(\alpha\), conducted on LLaVA-1.5-7B with a fixed block size of 2 and 192 retained tokens.
  This parameter balances text-conditional relevance (\(\alpha\) = 0.0) against intrinsic visual saliency (\(\alpha\) = 1.0).
  The optimal average performance is achieved at \(\alpha\)=0.8, showing the synergistic effect between the two components.}
  \label{tab:alpha_ablation}
  \begin{tabular}{c c | c c c c}
    \toprule
    \textbf{\(\alpha\) Value} & \textbf{Avg.(\%)} & \textbf{MME} & \textbf{MMB\textsuperscript{EN}} & \textbf{TextVQA} & \textbf{SQA\textsuperscript{IMG}} \\
    \midrule
    0.0 & 97.0\% & 1794 & 64.2 & 54.7 & 68.3 \\
    0.1 & 97.5\% & 1791 & 64.3 & 55.5 & 68.8 \\
    0.2 & 97.4\% & 1768 & 64.2 & 56.0 & 68.7 \\
    0.3 & 97.3\% & 1773 & 63.4 & 56.4 & 68.7 \\ 
    0.4 & 97.1\% & 1779 & 63.2 & 56.4 & 68.3 \\
    0.5 & 97.0\% & 1778 & 62.7 & 56.5 & 68.3 \\
    0.6 & 97.4\% & 1787 & 63.4 & 56.6 & 68.2 \\
    0.7 & 97.8\% & 1801 & 63.4 & 56.8 & 68.5 \\
    \textbf{0.8} & \textbf{98.1\%} & 1808 & 63.7 & 57.0 & 68.6 \\
    0.9 & 97.7\% & 1786 & 63.8 & 56.8 & 68.4 \\
    1.0 & 97.8\% & 1796 & 63.7 & 56.7 & 68.6 \\
    \bottomrule
  \end{tabular}
\end{table}

\subsection{Visualization}

To provide an intuitive understanding of the dynamics of our method, we visualize the token selection process of GridPrune in Fig. \ref{case}.
The figure shows that GridPrune is highly adaptive to user queries.
When asked to identify an object on the table (top row), the model focuses its limited token budget precisely on the calendar.
Similarly, when the query shifts to the object beside the dog (middle row) or a distant bird (bottom row), GridPrune dynamically reallocates its attention to the cup and the bird.
With an extremely sparse budget of only 12 tokens, the selection remains accurate, enabling the model to answer correctly.

This shows the advantage of GridPrune over instruction-irrelevant methods.
Rather than generating a generic summary of the image, our method performs task-driven active filtering of visual information.
This ability to form a targeted and relevant representation by preserving only the most important details is the key reason for its superior performance, allowing it to maintain better accuracy under high pruning conditions.

\section{Conclusion}

In this paper, we identify the limitation in existing visual token pruning methods: they primarily focus on the ``what to select" problem, while paying less attention to the ``where to look", which can lead to issues like inefficient spatial allocation, positional bias and so on.
So we propose GridPrune, a training-free method that divides pruning into a two-stage, ``guide-globally, select-locally" process.
GridPrune addresses the ``where to look" problem by using text-conditional guidance to dynamically allocate a token budget across spatial zones.
It then solves the ``what to select" problem through a localized selection within each budgeted zone.
Extensive experiments demonstrate that GridPrune achieves superior performance over state-of-the-art methods, particularly under aggressive pruning ratios.
Our work suggests that incorporating the ``where to look" problem into the pruning method is a promising direction for building more efficient and robust MLLMs.

{
    \small
    \bibliographystyle{ieeenat_fullname}
    \bibliography{main}
}


\end{document}